\documentclass[10pt,twocolumn,letterpaper]{article}

\usepackage[pagenumbers]{cvpr} 

\usepackage{graphicx}
\usepackage{amsmath}
\usepackage{amssymb}
\usepackage{booktabs}

\usepackage{pifont}
\usepackage[pagebackref,breaklinks,colorlinks]{hyperref}

\usepackage[capitalize]{cleveref}
\crefname{section}{Sec.}{Secs.}
\Crefname{section}{Section}{Sections}
\Crefname{table}{Table}{Tables}
\crefname{table}{Tab.}{Tabs.}

\def\cvprPaperID{2260} 
\def\confName{CVPR}
\def\confYear{2023}

\begin{document}

\title{One-Shot General Object Localization}

\author{Yang You\textsuperscript{1} \hspace{.1cm} Zhuochen Miao\textsuperscript{1,2} \hspace{.1cm} Kai Xiong\textsuperscript{1} \\ \hspace{.1cm} Weiming Wang\textsuperscript{1} \hspace{.1cm} Cewu Lu\textsuperscript{1} \\
\textsuperscript{1}Shanghai Jiao Tong University, China \hspace{.1cm}
\textsuperscript{2}	Tsinghua University, China \hspace{.1cm}
}
\maketitle

\begin{abstract}
   This paper presents a general one-shot object localization algorithm called OneLoc. Current one-shot object localization or detection methods either rely on a slow exhaustive feature matching process or lack the ability to generalize to novel objects. In contrast, our proposed OneLoc algorithm efficiently finds the object center and bounding box size by a special voting scheme. To keep our method scale-invariant, only unit center offset directions and relative sizes are estimated. A novel dense equalized voting module is proposed to better locate small texture-less objects. Experiments show that the proposed method achieves state-of-the-art overall performance on two datasets: OnePose dataset and LINEMOD dataset. In addition, our method can also achieve one-shot multi-instance detection and non-rigid object localization. Code repository: \href{https://github.com/qq456cvb/OneLoc}{https://github.com/qq456cvb/OneLoc}.
\end{abstract}

\section{Introduction}
\label{sec:intro}

Object detection has always been an important topic in computer vision. Many vision-related tasks such as human or object mesh reconstruction, human or object pose estimation rely heavily on the detection prior. However, current state-of-the-art object detection algorithms~\cite{ren2015faster,liu2016ssd,redmon2016you,carion2020end} require a large amount of annotated category-level training data (e.g., MS-COCO~\cite{lin2014microsoft}), and cannot generalize well to new instances which do not belong to any training category. In many real-world scenarios (e.g., robot manipulation), one-shot object detection is essentially useful, where a novel object need to be detected in a new environment, with only a few reference images of the object.

\begin{figure}[!ht]
    \centering
    \includegraphics[width=\linewidth]{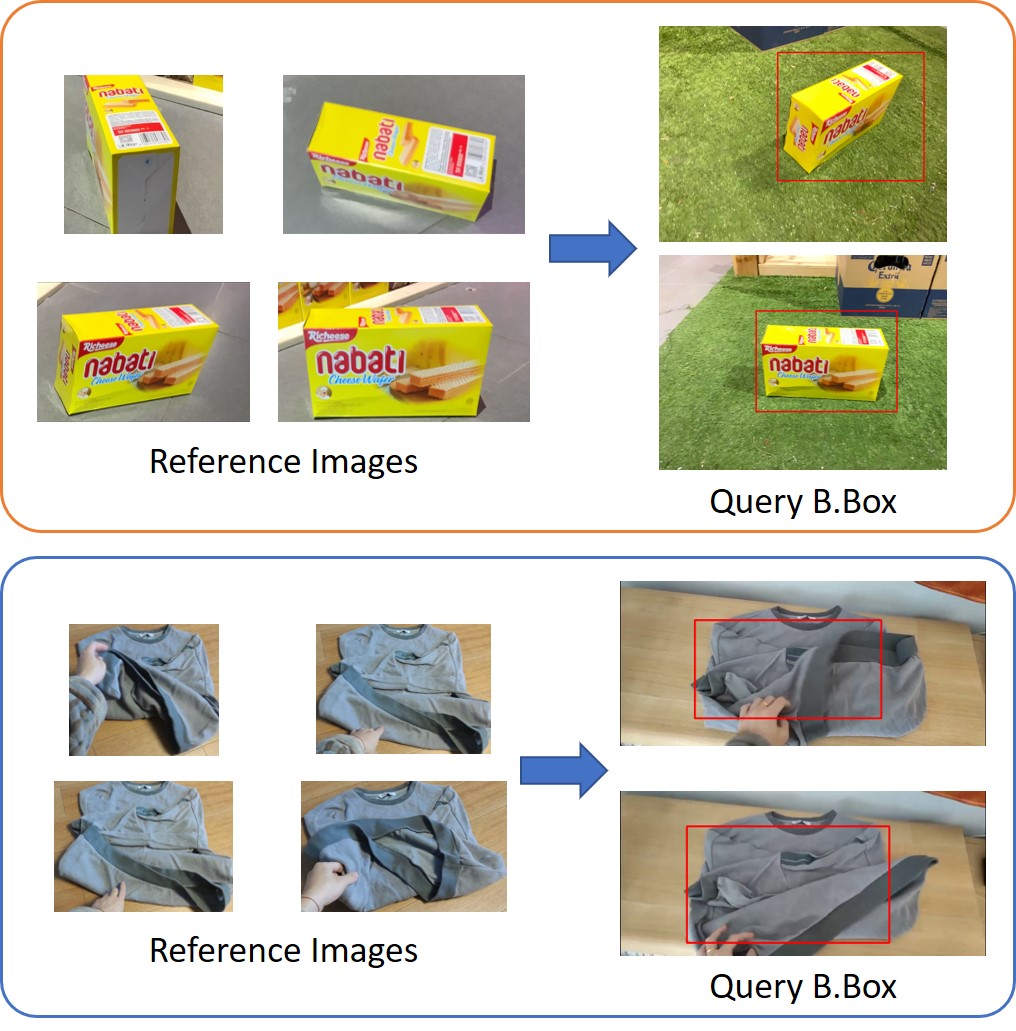}
    \caption{Given a few reference images, our method is able to accurately locate the target object in a novel query scene. The proposed method is general and cannot only detect rigid but also deformable objects.}
    \label{fig:intro}
\end{figure}

In this paper, we study the problem of one-shot object localization, a slightly simpler problem than object detection. In object localization, a target object is known to exist in the query image, and the objective is to locate this object, i.e., to give the 2D bounding box of this object. The term \textit{one-shot} means that Awe only have a reference video capturing this object during training. Under this problem setting, traditional methods~\cite{liu2017efficient,sattler2016efficient,toft2018semantic} usually create tons of object templates from different viewpoints and find the one that has the most matching features with the query image. However, this method is time-consuming and does not scale well in terms of template images. It also fails on texture-less small objects. Gen6D~\cite{liu2022gen6d} convolves the query image with kernels created from the reference images, and then outputs the object center with the largest response. However, it makes strong assumption of square bounding boxes and does not generalize well to novel datasets.

To overcome these challenges, we propose a novel \textbf{One}-shot \textbf{Loc}alization method called \textbf{OneLoc}. OneLoc is able to locate an object accurately given only a few reference images.
Given a reference image, we first sample a large collection of points in the image plane, and then compute their dense SuperPoint~\cite{detone2018superpoint} descriptors. For each point, we estimate its direction towards the bounding box center. We then pair these points randomly, and each point pair would cast a center vote analytically. The location that receives the most votes is considered the predicted center. For the estimation of the bounding box size, we additionally estimate the \textit{relative} size, which is the ratio between the center offset of the point and the absolute bounding box size. This strategy makes our model invariant to scale changes.

In addition, a dense voting algorithm with vote equalization is proposed in order to balance the number of votes in different areas. Instead of sampling sparse keypoints given by SuperPoint, we use \textit{stratified sampling} to sample a dense and equalized set of points in the image plane. Experiments show that this sampling strategy greatly improves the recall on those small, texture-less objects.

We evaluate our method on two public datasets, OnePose~\cite{sun2022onepose} and LINEMOD~\cite{hinterstoisser2012model}.  These two datasets are mostly used by previous works for pose estimation, given the bounding box of the target object. Since our goal is on one-shot object localization, we focus on the bounding box estimation of the target object, and report the recall of detected bounding boxes under different IoU thresholds. Results show that our method greatly outperforms previous algorithms and generalizes well to both datasets.

In addition, to explore the possibility of one-shot object \textit{detection}, we create a small detection dataset by capturing real-world videos containing objects from OnePose. In this dataset, multiple object instances may appear in the frame. Results show that our method is capable of accurately detecting multiple instances simultaneously. 

Our last experiment extends the proposed method to non-rigid object localization.
To summary, our contribution is included in the following three points:
\begin{itemize}
    \item We introduce a novel voting pipeline to achieve one-shot object localization. Pairwise center voting and relative size estimation are proposed to ensure accurate and scale-invariant bounding box prediction.
    \item A dense equalization vote sampling module is proposed to balance votes in different areas. This greatly improves the recall of small and texture-less objects.
    \item Experiments show that our method can not only locate but also detect multiple instances accurately. In addition, the proposed method is able to locate non-rigid objects in the wild.
\end{itemize}

\section{Related Work}
\subsection{Object Detection}
With the development of deep neural networks, numerous works have been proposed on general object detection. These methods assume a predefined list of target classes and train on a large collection of images with bounding box annotations (e.g., MSCOCO~\cite{lin2014microsoft}). Two-stage detectors such as Faster R-CNN~\cite{ren2015faster} and Mask R-CNN~\cite{he2017mask} use a region proposal network (RPN) and a detection head. They detect the object by sampling predefined anchors and then refine these anchors with a second-stage head. One stage detectors like YOLO~\cite{redmon2016you} and SSD~\cite{liu2016ssd} use only RPN-like networks and output bounding boxes directly in the corresponding hyperpixel. They are usually faster and less accurate than two-stage methods. The above methods, though pretty mature, cannot generalize to novel classes or instances that do not appear in the training set.

\subsection{One-Shot Object Localization and Detection}
Recently, a few works have explored one-shot object localization/detection.
HF-Net~\cite{sarlin2019coarse} proposes a hierarchical localization framework, where a set of reference images are first searched based on the global descriptors, and then fine-grained feature matching is conducted to find the final estimated pose. Gubbi \textit{et al.}~\cite{venkatesh2019one} use fully connected siamese networks to locate the object with the highest response. However, their method can only handle simple objects with clean backgrounds, with limited generalization ability. Besides deep learning based localization, there are also a set of traditional methods~\cite{liu2017efficient,sattler2016efficient,toft2018semantic} that conduct exhaustive matching between the query image and a 3D SfM model database. These methods, though more accurate, are compute-intensive and not scalable in terms of the number of template images.

For one-shot object detection, OS2D~\cite{osokin2020os2d} builds a one-stage system that performs localization and recognition jointly, where dense correlation matching of learned local features is computed to find correspondences. However, it can handle only a single template image without viewpoint changes. DTOID~\cite{mercier2021deep} is the current state-of-the-art one-shot detection method. It designs an architecture that computes the correlation between the query image and template images, and then regresses bounding boxes with a separate head. This method, however, achieves mediocre performance on unseen object instances (e.g., objects in OnePose dataset). OSSID~\cite{gu2022ossid} proposes a self-supervised learning pipeline to fine-tune the performance of the model in the test data. It requires the CAD model of the target object and depth input in order to train the model. TDID~\cite{ammirato2018target} and Gen6D~\cite{liu2022gen6d} both use correlation between templates and queries to locate the target object. However, they do not generalize well to novel objects.

\begin{figure*}[ht]
    \centering
    \includegraphics[width=0.9\linewidth]{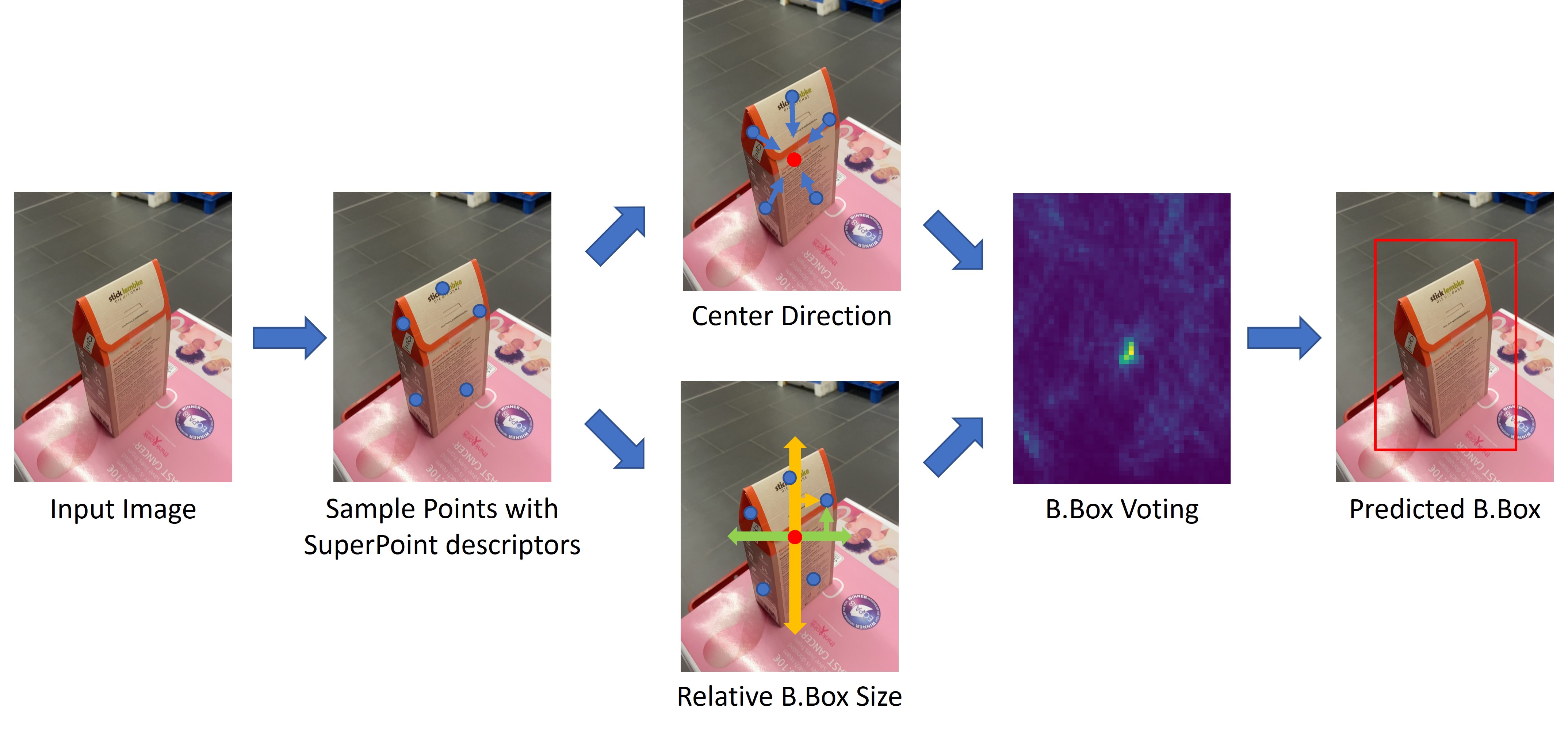}
    \caption{\textbf{An illustration of our model pipeline.} Taking an input image, we sample a set of points across the image plane. Then the interpolated SuperPoint descriptors at these point locations are fed into a neural network to predict the direction towards the bounding box center and the relative size of the bounding box. A dense voting scheme is leveraged to output the final bounding box.}
    \label{fig:pipeline}
\end{figure*}

\section{Method}

Our method assumes that a set of reference images containing the target instance are provided. In all our experiments, these reference images are extracted from a single reference video. The goal of our method is to rapidly and accurately detect the target object's bounding box in a query frame given the set of reference images.

Our method first takes an input query image, and then sample a set of points together with their SuperPoint~\cite{detone2018superpoint} descriptors. For each sampled point, our network predicts a unit offset direction towards the bounding box center. At test time, we randomly sample a large number of point pairs. For each point pair, together with their predicted unit offsets, the center of the bounding box can be computed analytically and cast a vote in the corresponding location. The location that receives the most votes is considered the predicted center. In order to get the bounding box size, we additionally estimate a relative bounding box size for each sampled point, and accumulate them during the center voting process. We also propose a novel dense voting scheme with equalization to balance the votes in different regions.

The proposed localization method can be extended to one-shot detection, where a Non-Maximum-Suppression is conducted on the center vote map. An overview of our method is illustrated in Figure~\ref{fig:pipeline}.

\subsection{Learning to Vote for Object Center}
Denote the image as $\mathbf{I}\in\mathbb{R}^{H\times W}$, we sample a set of points $\{\mathbf{p}|\mathbf{p}\in\mathbb{R}^2\}$ within the bounding box and then predict a unit offset towards the bounding box center in the image plane. Considering a point $\mathbf{p}$ and the center of the annotated box $\mathbf{o}$, our neural network will take its SuperPoint descriptor as input and output the unit offset direction $\mathbf{d}(\mathbf{p})$ given by:

\begin{align}
    \mathbf{d}(\mathbf{p}) = \frac{\mathbf{p} - \mathbf{o}}{\|\mathbf{p} - \mathbf{o}\|_2}.
\end{align}

It is important to use the direction instead of absolute offsets, in order to make our algorithm scale-invariant. As we only have a set of reference images at training time in our problem setting, the scale of the target object is usually unknown during testing. 


During inference, inspired by PVNet~\cite{peng2019pvnet}, we randomly sample point pairs and cast a center vote for each pair. Specifically, given two offset directions $\mathbf{d}(\mathbf{p}_1), \mathbf{d}(\mathbf{p}_2)$ and their starting points $\mathbf{p}_1, \mathbf{p}_2$, the object center $\mathbf{c}$ can be analytically computed by the intersection of the two:
\begin{align}
    \mathbf{c} =  \mathbf{p}_1 + \frac{(\mathbf{p}_2 - \mathbf{p}_1) \times \mathbf{d}(\mathbf{p}_2)}{\mathbf{d}(\mathbf{p}_1)\times \mathbf{d}(\mathbf{p}_2)}\cdot \mathbf{d}(\mathbf{p}_1),
\end{align}

where $\times$ is the cross product, we also discard the case where $\frac{(\mathbf{p}_2 - \mathbf{p}_1) \times \mathbf{d}(\mathbf{p}_2)}{\mathbf{d}(\mathbf{p}_1)\times \mathbf{d}(\mathbf{p}_2)} < 0$, which means that the intersection is out of the ray $(\mathbf{p}_1, \mathbf{d}(\mathbf{p}_1))$.

\paragraph{Voting for Center or Corner?}
The choice to vote for object bounding box center is not arbitrary. One may also think of voting for the top-left/bottom-right 2D object bounding box corners to predict the object center and scale all together. However, the variance of corner votes is much larger than center, which is also observed in previous work~\cite{peng2019pvnet}.

\begin{figure}[ht]
    \centering
    \includegraphics[width=0.9\linewidth]{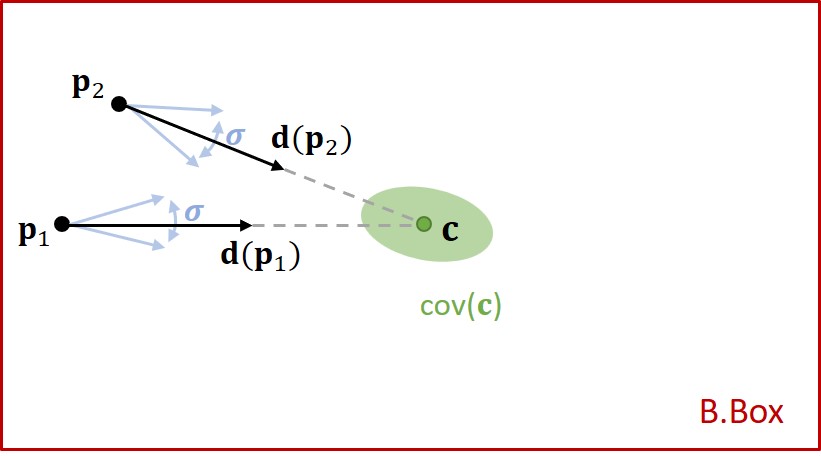}
    \caption{Given a pair of points $\mathbf{p}_1,\mathbf{p}_2$, the bounding box center $\mathbf{c}$ can be uniquely determined. We assume that the predicted direction towards $\mathbf{c}$ has a variance of $\sigma$, which results in a covariance matrix in the predicted $\mathbf{c}$.}
    \label{fig:centervote}
\end{figure}
To analyze this behavior rigorously, without loss of generality, suppose $\mathbf{p}_1=(0, 0), \mathbf{p}_2=(a, b)$ and $\mathbf{d}(\mathbf{p}_1)=(\cos\alpha, \sin\alpha), \mathbf{d}(\mathbf{p}_2)=(\cos\beta, \sin\beta)$. We further assume that offset directions $\alpha,\beta$ have a Gaussian prediction noise with standard deviation $\sigma$, as illustrated in Figure~\ref{fig:centervote}. The center vote $\mathbf{c}$ is now a function of the predicted offset directions $\mathbf{c}=f(\alpha,\beta)$. Then according to the Result 5.6 of ~\cite{hartley2003multiple}, the covariance of center vote $\mathbf{c}\in\mathbb{R}^2$ is:

\begin{align}
\label{eq:cov}
    \mathrm{cov}(\mathbf{c}) = \mathrm{J} \Sigma \mathrm{J}^T,
\end{align}
where $\mathrm{J}\in\mathbb{R}^{2\times 2}$ is the Jacobian matrix of the function $f$ and $\Sigma\in\mathbb{R}^{2\times 2}$ is the diagonal matrix with diagonal elements set to $\sigma^2$.

\begin{figure}[h]
    \centering
    \includegraphics[width=\linewidth]{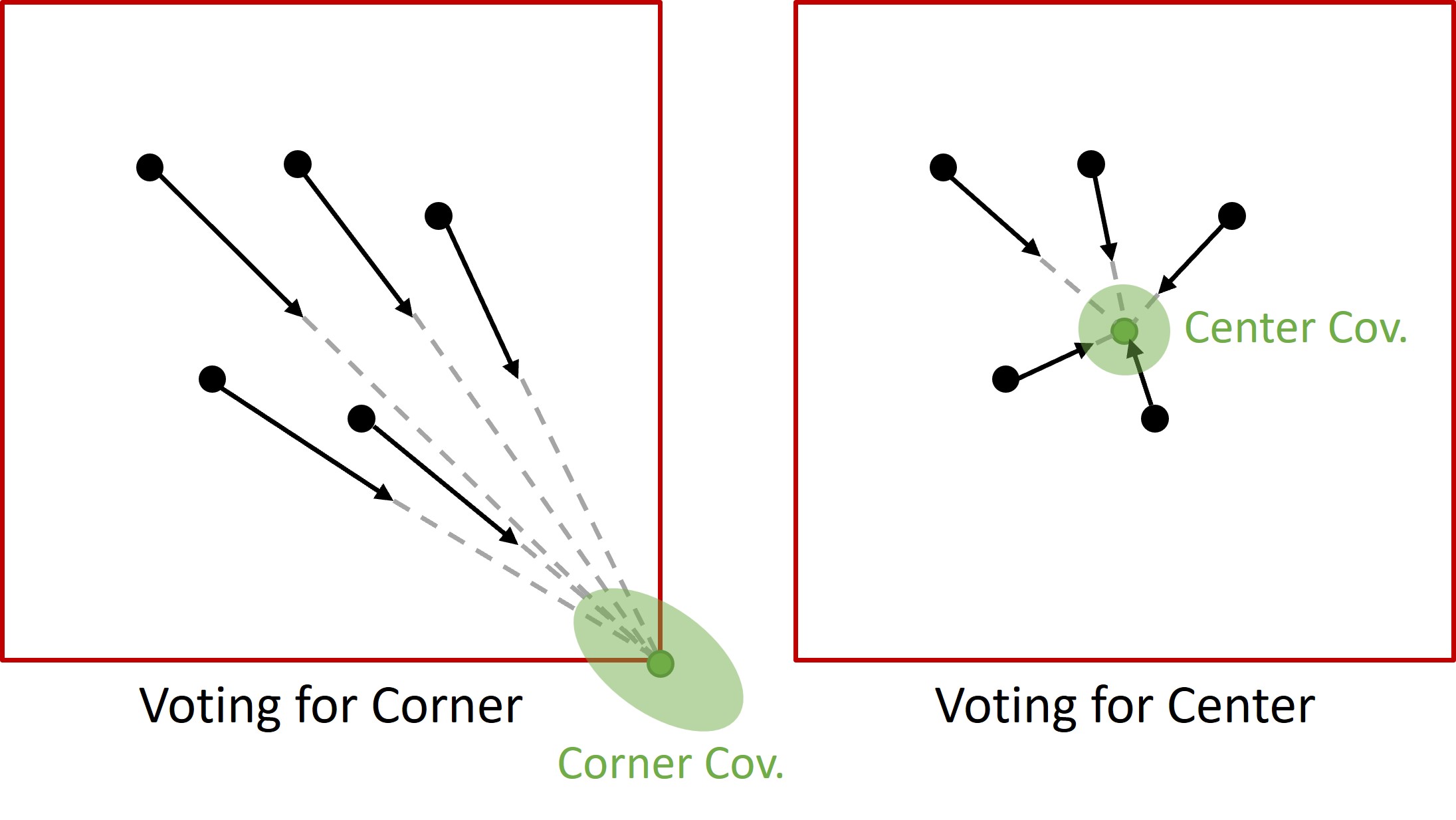}
    \caption{\textbf{Comparison of voting for corners and centers.} The variance of voting for corners is much larger than voting for centers.}
    \label{fig:corner}
\end{figure}

To compute $\mathrm{cov}(\mathbf{c})$, we substitute $\alpha,\beta$ into $J$, and without loss of generality, let $\alpha=0$, we have:
\begin{align}
    \mathrm{det}(\mathrm{J})=\frac{b\cdot(b\cos\beta - a\sin\beta)}{\sin^3\beta},
\end{align}
and the determinant of the final covariance matrix $\mathbf{c}$ can be computed as:
\begin{align}
    \mathrm{det}(\mathrm{cov}(\mathbf{c})) &=\sigma^4\mathrm{det}^2(\mathrm{J}),
\end{align}
with detailed deduction given in the supplementary. Therefore, 
if we fix $a, b$, the angle $\beta$ between the two unit directions $\mathbf{d}(\mathbf{p}_1), \mathbf{d}(\mathbf{p}_2)$ is nearly inverse proportional to $\mathrm{det(J)}$, and as $\beta\rightarrow 0$, $\mathrm{det(J)}\rightarrow \infty$. As shown in Figure~\ref{fig:corner}, if we vote for bounding box corners, the predicted directions towards the corner all lie within $[-90^\circ, 0]$. Sampling point pairs from this set is more likely to have small relative angle $\beta$ than voting for centers, where the predicted directions lie in the full $[0, 360^\circ]$.

\subsection{Learning to Vote for Object Size}
In order to locate the 2D bounding box, we need to predict the bounding box size besides the box center. Likewise, we also generate a size vote $\mathbf{s}(\mathbf{p})\in \mathbb{R}^2$ for each point. This size vote is computed by the ratio between the point's center offset and the absolute size of the bounding box:

\begin{align}
\label{eq:rel_size}
    \mathbf{s}(\mathbf{p}) = \frac{\mathbf{p} - \mathbf{c}}{\mathbf{\bar{s}}},
\end{align}
where $\mathbf{\bar{s}}\in\mathbb{R}^2$ is the ground-truth bounding box size and the division is element-wise.

\paragraph{Relative Size v.s. Absolute Size.}
Instead of estimating the absolute box size $\mathbf{\bar{s}}$, we use the relative size in Equation~\ref{eq:rel_size}. This is important, since regressing the relative size keeps our model scale-invariant. For objects that may appear with different box sizes in test images, the relative size for point $\mathbf{p}$ is the same and the SuperPoint descriptor for this point is also similar (because SuperPoint is scale-invariant~\cite{detone2018superpoint,sarlin2019coarse}). Therefore, the network will not get confused about the relationship between inputs and outputs. If, in contrast, we force the model to regress different absolute sizes for similar input descriptors, the network has to learn an improper relationship and diverges.

At test time, we cast a vote for each sampled point and compute the average size in the predicted box center.

\begin{figure*}[ht]
    \centering
    \includegraphics[width=0.85\linewidth]{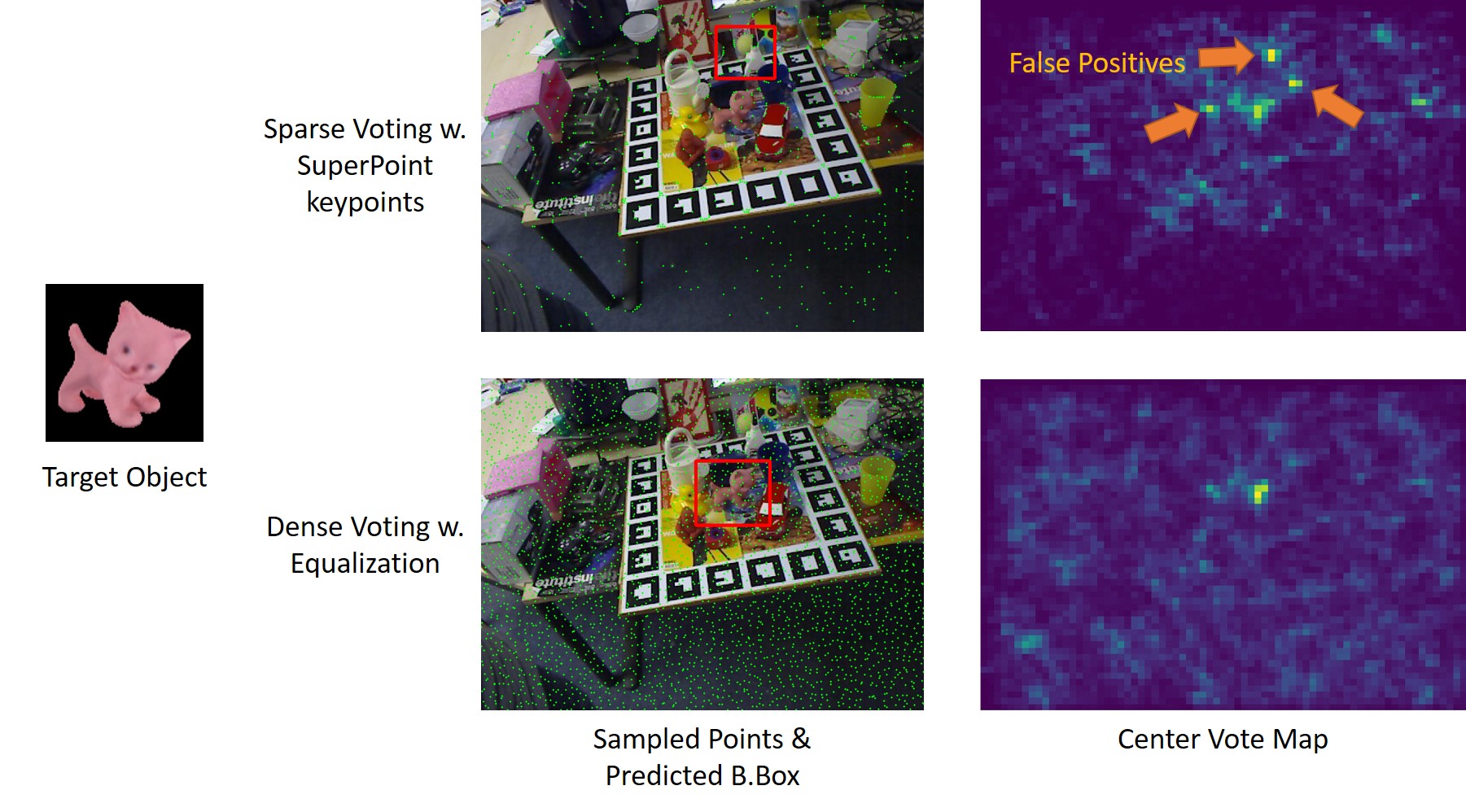}
    \caption{\textbf{Dense voting with equalization.} We compare the proposed dense voting module with sparse voting. If we vote with detected SuperPoint keypoints (shown in green), false positives would appear at those locations with more keypoints detected. In contrast, our proposed dense equalized voting scheme balances the point density across different areas and gives a better result (shown in red b.box).}
    \label{fig:dense}
\end{figure*}
\subsection{Dense Vote Sampling with Equalization}
The above sections discussed the voting scheme given a collection of sampled points $\{\mathbf{p}|\mathbf{p}\in\mathbb{R}^2\}$, but the sampling strategy itself is also important and can greatly influence the final performance.

SuperPoint gives a dense keypoint probability map $\Omega\in\mathbb{R}^{H\times W}$ and a dense descriptor map $\Phi\in\mathbb{R}^{H\times W\times 256}$ given the query image $\mathbf{I}\in\mathbb{R}^{H\times W}$. In practice, one usually samples a sparse set of keypoints by applying Non-Maximum-Suppression (NMS) on the probability map $\Omega$. Though these keypoints do generalize across different scenes. They are usually sparse, especially for those low-textured objects, as shown in Figure~\ref{fig:dense}. There are fewer votes in these areas and the target object is more likely to be missed. As a result, we cannot say anything about the predicted center, since it may be a false positive just because there are more keypoints detected around this center. 

To solve this problem, we propose a dense voting module with vote equalization. Specifically, we discard the keypoint generation process in SuperPoint, but instead sample points densely across the entire image. The descriptors of these sampled points are interpolated from the dense descriptor map.
Stratified sampling is leveraged in order to balance the number of sampled points in different areas. Specifically,
the entire 2D image plane is divided into $M\times N$ square grids/strata. Within each grid/stratum, a random point is sampled by jittering the center point of the stratum by a random offset up to half the stratum’s width and height. A comparison of sampled points and center voting heatmaps between ours and SuperPoint is given in Figure~\ref{fig:dense}. This dense voting technique greatly improves the performance, especially on those texture-less objects.

\subsection{Implementation Details}
Our network is composed of 20 hidden residual-like~\cite{he2016deep} fully connected layers, with hidden size 128. The network gives four outputs, the first two is the unit center offset, and the last two is the relative bounding box size. The offset prediction is trained by the negative cosine similarity loss, powered by two, while the relative bounding box is supervised by the mean squared loss. The model is trained with Adam optimizer, with learning rate 3e-4 and weight decay 1e-4, for 50 epochs. The batch size is 4. We apply Gaussian blur, Gaussian noise, ISO noise, and brightness/contrast data augmentation during training.

In vote equalization, the grid or strata size is computed by the smaller side of the image divided by 50. Besides, to reduce GPU memory consumption, the center voting accumulation grid is also down-sampled by the same size. For point pair sampling, we limit the distance between point pairs not larger than 1/4 and 1/8 of the image size, for OnePose and LINEMOD, respectively. We also scale all the query images to $640\times 480$ or $480\times 640$ for portraits.

\section{Experiments}
In this section, we evaluate the performance of the proposed method in three tasks: one-shot rigid object localization, one-shot rigid object detection, and one-shot non-rigid object localization.

\subsection{One-Shot Rigid Object Localization}
\paragraph{Datasets.} We evaluate our method on two public datasets: OnePose~\cite{sun2022onepose} dataset and LINEMOD~\cite{hinterstoisser2012model} dataset. OnePose dataset contains more than 450 video sequences of 150 objects. For each object, multiple video recordings, accompanied camera poses, and 3D bounding box annotations are provided. These sequences are collected under different background environments, and each has an average recording-length of 30 seconds, covering all views of the object. We follow OnePose to evaluate on the 80 test objects. For each object, a reference sequence is used for training and a test sequence is used for evaluation. The test video sequence is sampled every 5 frames to maximize the visual difference. LINEMOD dataset is a widely used dataset for object pose estimation. On the LINEMOD dataset, we follow Gen6D~\cite{liu2022gen6d} to evaluate on five objects: \textit{cat}, \textit{duck}, \textit{bench vise}, \textit{cam}, and \textit{driller}. We also follow the split of ~\cite{tekin2018real}, and select the training images (about 180) as reference images and all the rest (about 1000) test images as query images for each object.

\paragraph{Evaluation Metrics.}
Since the objective of this paper is to accurately locate the 2D bounding box of the target object, we compute the recall of detected 2D bounding boxes with regard to the ground-truth under different IoU ratios. Mean recalls (mRec$_{25}$, mRec$_{50}$) over all objects are reported. The ground-truth 2D bounding box in OnePose dataset is generated by the tightest bounding box of the projected 3D bounding box. For a fair comparison, the ground-truth bounding box in LINEMOD dataset is generated by the tightest bounding box of a unit sphere, to align with the baseline proposed in Gen6D~\cite{liu2022gen6d}. 

In addition, to show that the proposed method is useful in downstream tasks, i.e., 3D pose estimations, we also evaluate the quality of estimated object pose given the predicted 2D bounding box. Specifically, for OnePose dataset, we crop the image based on the 2D bounding box, and then feed the cropped image into OnePose pose estimator to get the predicted 3D pose; for LINEMOD dataset, we crop the image likewise, and then feed the cropped image into Gen6D pose estimator to get the predicted pose. For a fair and consistent evaluation, we use the \textit{5cm-5deg} metric proposed in ~\cite{sun2022onepose} on OnePose dataset. \textit{5cm-5deg} metric deems a predicted pose correct if the error is below 5 cm and 5 degrees. ADD-0.1d metric is used on LINEMOD dataset,  which computes the recall rate of ADD~\cite{hinterstoisser2012model} with 10\% of the object diameter. We also compute the recall rate at 5 pixels (Prj-5), following Gen6D.

\paragraph{Baselines.}
We compare our method with three baselines: traditional feature matching, Gen6D detector~\cite{liu2022gen6d} and DTOID~\cite{mercier2021deep}. The feature matching method first computes SuperPoint~\cite{detone2018superpoint} features of the reference images as a database. Given a query image, it exhaustively matches the image with all references using SuperGlue~\cite{sarlin2020superglue}, and the reference image with the largest number of inliers is considered the final match. The predicted 2D bounding box is given by warping the reference image corners with the estimated affine transform. Gen6D is a generalizable pose estimator, and we use its detector component as a baseline. Gen6D detector convolves the query image feature with the reference image features and outputs a score map and a scale map. The location with the highest score is considered the center of the predicted bounding box center, and the scale is also given in this location. DTOID computes the correlation between the query image and template images, and then regresses bounding boxes with a separate head.

\paragraph{Results.}

\begin{table}[ht]
    \centering
    \resizebox{\linewidth}{!}{
    \begin{tabular}{lcccc}
    \toprule
     & \multicolumn{2}{c}{OnePose} & \multicolumn{2}{c}{LINEMOD}\\
     \cmidrule(lr){2-3}\cmidrule(lr){4-5}
     & mRec$_{25}$ & mRec$_{50}$ & mRec$_{25}$ & mRec$_{50}$ \\
     \midrule
     Feat. Match~\cite{detone2018superpoint,sarlin2020superglue} & \textcolor{red}{97.60} & \textcolor{red}{77.95} & 40.28 & 31.43 \\
      Gen6D~\cite{liu2022gen6d} & 24.97 & 2.58 & \textcolor{red}{95.90} & \textcolor{red}{93.99} \\
     DTOID~\cite{mercier2021deep} & 66.84 & 32.18 & 57.61 & 44.19 \\
     \midrule
     Ours & \textcolor{blue}{87.80} & \textcolor{blue}{73.39} & \textcolor{blue}{89.80} & \textcolor{blue}{79.31} \\
     \bottomrule
    \end{tabular}
    }
    \caption{\textbf{Performance of 2D bounding box localization on OnePose and LINEMOD dataset.} The best is shown in red and the second best is shown in blue. Our method gives the best overall performance.}
    \label{tab:loc}
\end{table}

\begin{table}[ht]
    \centering
    \begin{tabular}{lccc}
    \toprule
     & OnePose & \multicolumn{2}{c}{LINEMOD}\\
     \cmidrule(lr){2-2}\cmidrule(lr){3-4}
     & 5cm-5deg & ADD-0.1d & Prj-5 \\
     \midrule
     Feat. Match~\cite{detone2018superpoint,sarlin2020superglue} & \textcolor{red}{81.29} & 24.73 & 26.78 \\
     Gen6D~\cite{liu2022gen6d} & 27.01 & \textcolor{red}{62.45} & \textcolor{red}{84.37} \\
     DTOID~\cite{mercier2021deep} & 60.08 & 22.81 & 33.71 \\
     \midrule
     Ours & \textcolor{blue}{74.52} & \textcolor{blue}{55.55} & \textcolor{blue}{77.02} \\
     \bottomrule
    \end{tabular}
    \caption{\textbf{Performance of pose estimation using predicted bounding boxes on OnePose and LINEMOD dataset.} The best is shown in red and the second best is shown in blue. Our method gives the best overall performance.}
    \label{tab:pose}
\end{table}

\begin{figure*}[ht]
    \centering
    \includegraphics[width=\linewidth]{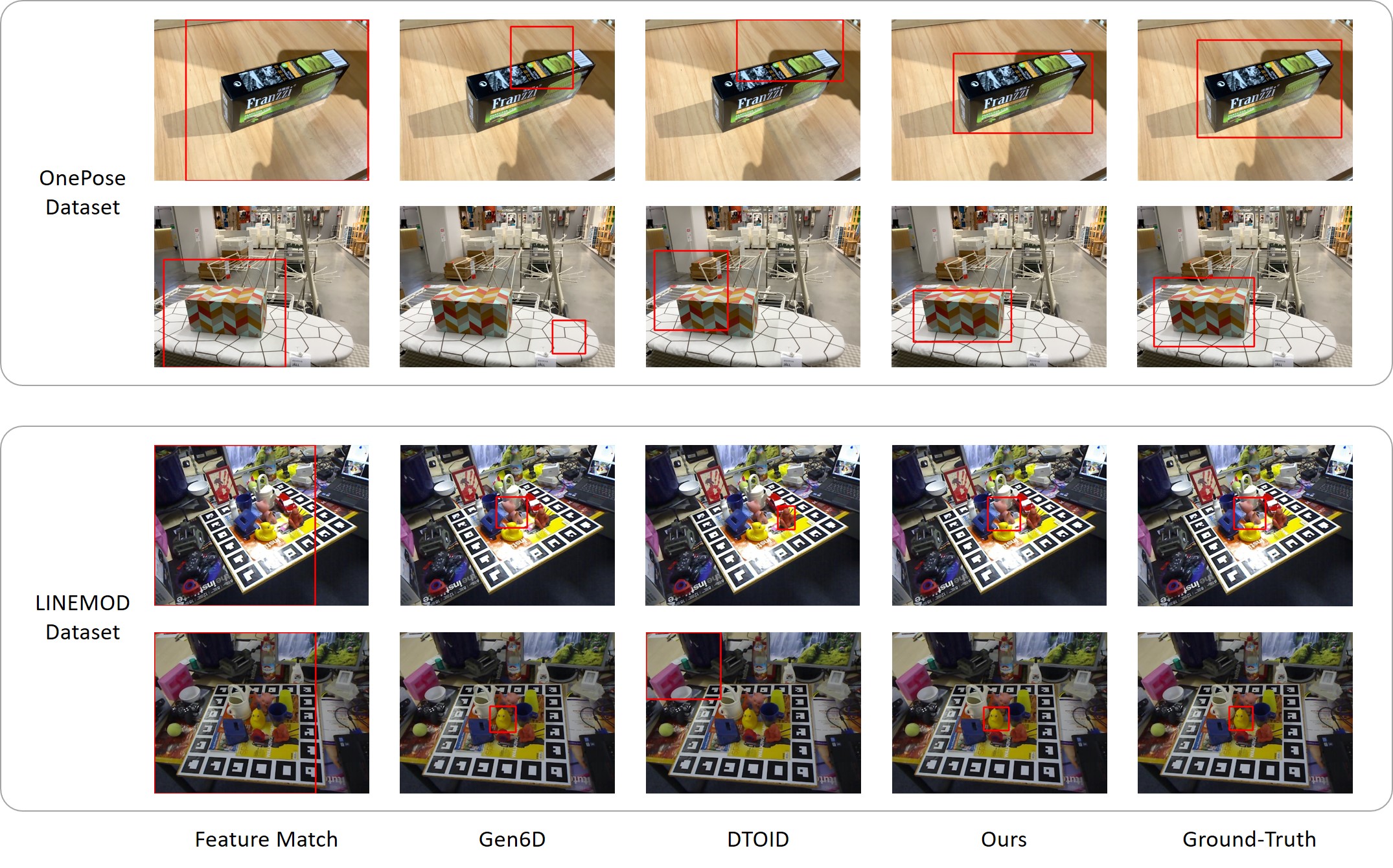}
    \caption{\textbf{Qualitative results on OnePose and LINEMOD dataset. }Our method is generalizable across both datasets. }
    \label{fig:vis}
\end{figure*}
Quantitative results are given in Table~\ref{tab:loc} and \ref{tab:pose}. Previous methods significantly overfit one dataset and fail on the other. For example, Gen6D, though performs well on LINEMOD dataset, it fails to detect the larger bounding box on OnePose dataset. The feature matching method performs well on OnePose dataset, but fails to give reasonable performance on LINEMOD dataset, which contains small texture-less objects. Besides, DTOID only achieves mediocre performance on these two datasets and its ability of generalization is limited. In contrast, our method performs the best in terms of the average performance on both datasets. It outperforms the best baseline by \textbf{19.88} mRec$_{25}$ and \textbf{21.66} mRec$_{50}$. Some qualitative results are shown in Figure~\ref{fig:vis}.

\subsection{One-Shot Rigid Object Detection}
In this experiment, we extend our method to one-shot object detection. This is achieved by simply applying Non-Maximum-Suppression (NMS) operation on the resulting vote map. This task is more difficult than object localization, since there may be more than one instance in the query image.

\paragraph{Datasets.} To the best of our knowledge, currently there is no mature one-shot detection benchmark with multiple instances. Therefore, we collect and annotate three video sequences containing the three objects from OnePose dataset. Each sequence is around 15 seconds long, with a sampling rate of every 10 frames for evaluation.

\paragraph{Evaluation Metrics.} We use the common average precision metric in the evaluation of object detection with respect to different IoU ratios, abbreviated by AP$_{25}$ and AP$_{50}$.

\begin{table}[h]
    \centering
    \resizebox{\linewidth}{!}{
    \begin{tabular}{lcccccc}
    \toprule
    & \multicolumn{2}{c}{Sequence 1} & \multicolumn{2}{c}{Sequence 2} & \multicolumn{2}{c}{Sequence 3}\\
    \cmidrule(lr){2-3}\cmidrule(lr){4-5}\cmidrule(lr){6-7}
    & AP$_{25}$ & AP$_{50}$ & AP$_{25}$ & AP$_{50}$ & AP$_{25}$ & AP$_{50}$\\
     \midrule
     DTOID~\cite{mercier2021deep} & 65.22 & 65.22 & 56.20 & 20.53 & 54.16 & 52.75 \\
     \midrule
     Ours & \textbf{73.11} & \textbf{72.03} & \textbf{83.65} & \textbf{40.46} & \textbf{79.59} & \textbf{67.74} \\
     \bottomrule
    \end{tabular}
    }
    \caption{\textbf{Multi-instance detection results.}}
    \label{tab:detection}
\end{table}

\begin{figure}[h]
    \centering
    \includegraphics[width=\linewidth]{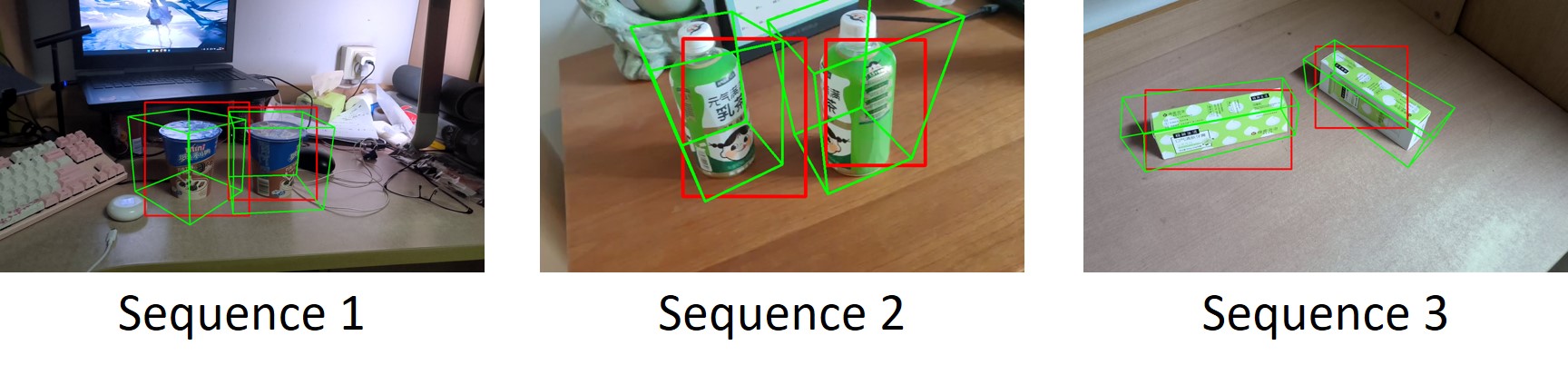}
    \caption{\textbf{Visualization of successfully predicted bounding boxes (red) and poses (green) on the three sequences.}}
    \label{fig:multi}
\end{figure}

\paragraph{Results.} Quantitative results are given in Table~\ref{tab:detection}. We also compare our method with DTOID. The proposed method consistently outperforms DTOID in all three sequences. Some qualitative results are also shown in Figure~\ref{fig:multi}.

\subsection{Ablation Studies and Analysis}

\begin{table}[h]
    \centering
    \begin{tabular}{lcccc}
    \toprule
     & \multicolumn{2}{c}{OnePose} & \multicolumn{2}{c}{LINEMOD}\\
     \cmidrule(lr){2-3}\cmidrule(lr){4-5}
     & mRec$_{25}$ & mRec$_{50}$ & mRec$_{25}$ & mRec$_{50}$ \\
     \midrule
     Corner Voting & 18.58 & 8.11 & 7.58 & 0.43 \\
     Sparse Voting & 78.73 & 61.41 & 42.27 & 30.60 \\
     \midrule
     Ours & \textbf{87.80} & \textbf{73.39} & \textbf{89.80} & \textbf{79.31} \\
     \bottomrule
    \end{tabular}
    \caption{\textbf{Ablation study results.}}
    \label{tab:ablation}
\end{table}

\paragraph{Voting Centers v.s. Voting Corners.}
As we previously discussed, voting corners results in a large variance compared to voting centers, making it harder to locate the peak. This is also validated in Table~\ref{tab:ablation}. Voting corners gives much worse results than voting centers.

\paragraph{Relative v.s. Absolute Size Estimation.}
Another important strategy is to regress the relative size rather than the absolute. To show why, we resize the original OnePose dataset images into different sizes, to simulate the object appearing at different scales. The results for images of different sizes are given in Table~\ref{tab:abs_size}. Though regressing absolute size achieves slightly better results on the original dataset, this dataset bias is broken when the query image is resized. Performance drops significantly on $384\time 288$ and $960\times 720$, especially for mRec$_{50}$. On the contrary, our method with relative size estimation, though completely trained on $640\times 480$ images, generalizes well across different scales.

\begin{table}[h]
    \centering
    \resizebox{\linewidth}{!}{
    \begin{tabular}{lcccccc}
    \toprule
     & \multicolumn{2}{c}{$640\times 480$ } & \multicolumn{2}{c}{$384\times 288$} & \multicolumn{2}{c}{$960\times 720$}\\
     \cmidrule(lr){2-3}\cmidrule(lr){4-5}\cmidrule(lr){6-7}
     & mRec$_{25}$ & mRec$_{50}$ & mRec$_{25}$ & mRec$_{50}$ & mRec$_{25}$ & mRec$_{50}$\\
     \midrule
     Abs. Size & \textbf{89.57} & \textbf{78.84} & \textbf{83.46} & 35.96 & 79.58 & 44.93 \\
     \midrule
     Ours & 87.80 & 73.39 & 82.20 & \textbf{65.80} & \textbf{80.30} & \textbf{58.56} \\
     \bottomrule
    \end{tabular}
    }
    \caption{\textbf{Ablation study on absolute size estimation.}}
    \label{tab:abs_size}
\end{table}

\paragraph{Sparse v.s. Dense Vote Sampling.}
As we mentioned before, if we only use the sparse SuperPoint keypoints to vote, texture-less and small objects are more likely to get missed. This is also confirmed in Table~\ref{tab:ablation}, where the performance drop on LINEMOD dataset is more significant than on OnePose dataset, since LINEMOD dataset contains more texture-less objects like \textit{duck} or \textit{cat}.

\paragraph{Training and Running Time.}
Although our method needs to be trained on reference images, this training process is extremely fast and can be completed in 15 minutes. This allows a fast deployment of the proposed method, where one can capture a reference video of the target object and obtain a generalizable object detector in 15 minutes.

In addition, our method takes constant run-time regardless of the number of reference templates and target instances in the query image. In contrast, the run-time of previous baselines is linear with regard to the number of templates and may require multiple forward passes if there is more than one instance in the query image. Quantitatively, the proposed method takes an average of 133ms per image on a single 1080Ti GPU. In contrast, feature matching and DTOID take more than 2s to process a query image with 200+ templates in the database.


\subsection{Non-Rigid Object Localization}
\begin{figure}[h]
    \centering
    \includegraphics[width=\linewidth]{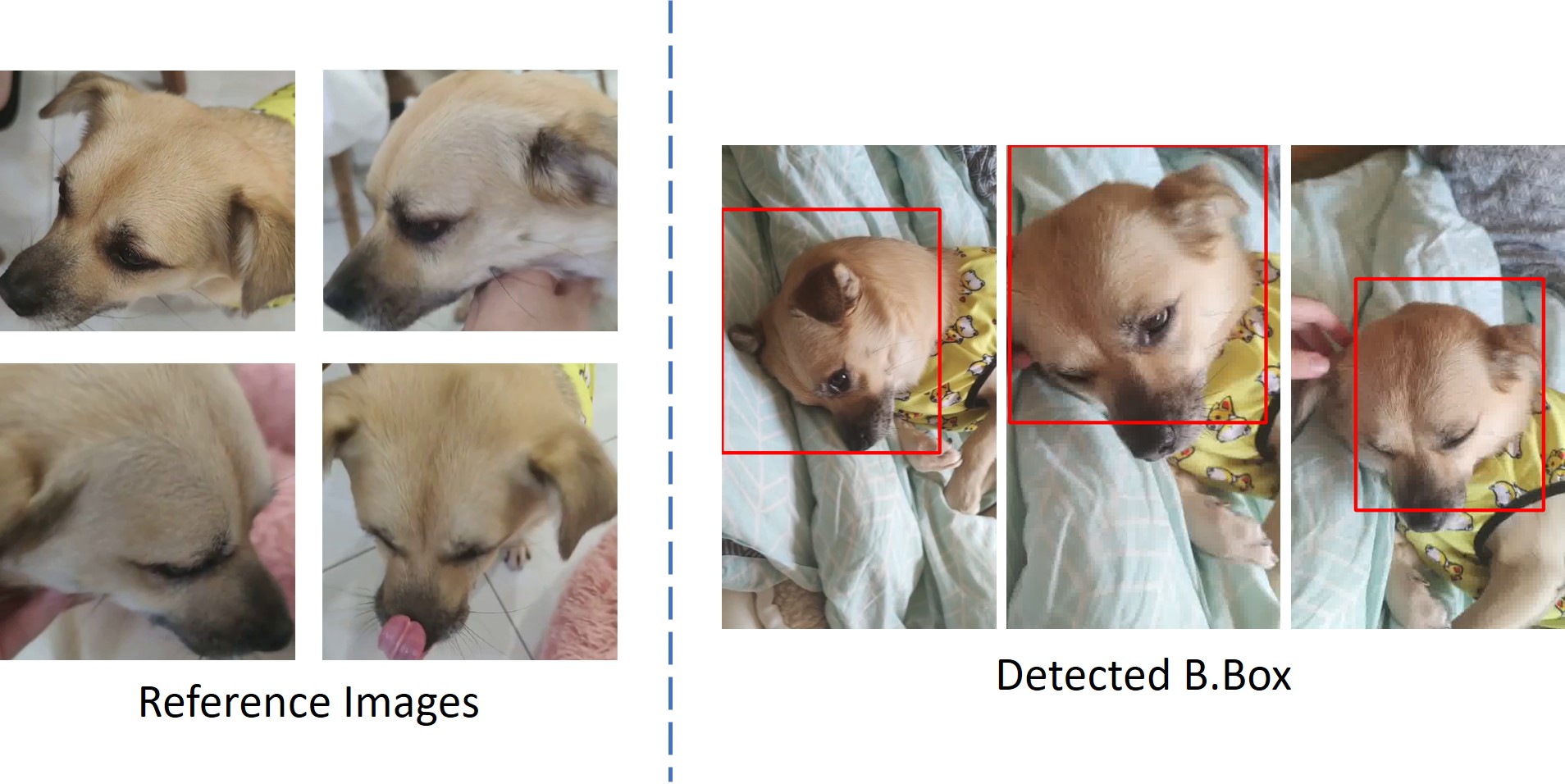}

    \includegraphics[width=0.9\linewidth]{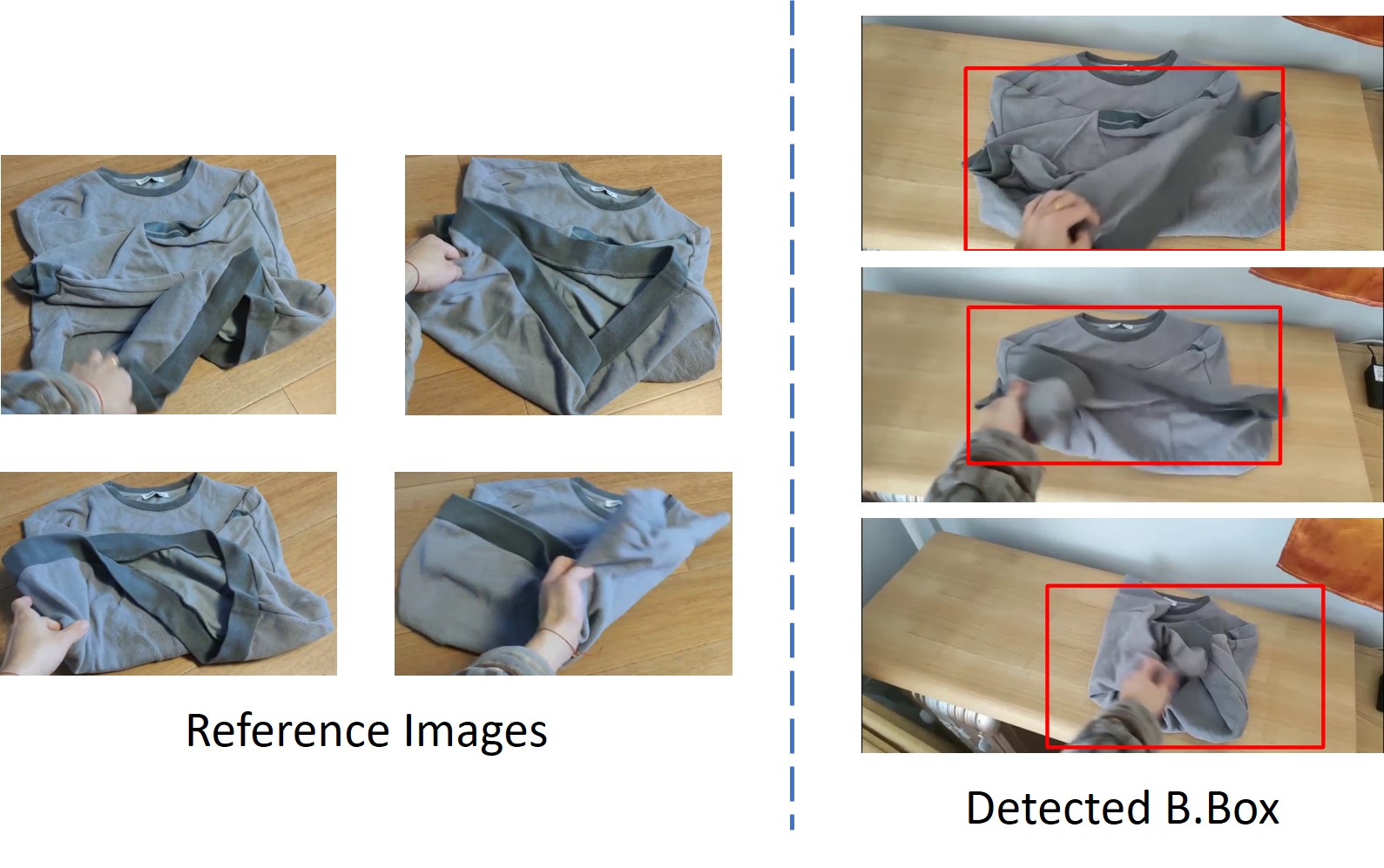}
    \caption{\textbf{Qualitative results on non-rigid object localization.}}
    \label{fig:nonrigid}
\end{figure}

Our method cannot only localize rigid objects but also non-rigid objects, given only a few reference images of the target object. In this section, we collect and annotate two sequences of non-rigid objects: \textit{dog} and \textit{clothes}. For each object, a reference sequence is used for training while the other is for testing. The test sequence is sampled every 10 frames.  Results are given in Table~\ref{tab:nonrigid} and Figure~\ref{fig:nonrigid}.

\begin{table}[h]
    \centering
    \begin{tabular}{lcccc}
    \toprule
    & \multicolumn{2}{c}{Dog} & \multicolumn{2}{c}{Clothes}\\
    \cmidrule(lr){2-3}\cmidrule(lr){4-5}
    & Rec$_{25}$ & Rec$_{50}$ & Rec$_{25}$ & Rec$_{50}$\\
     \midrule
     DTOID~\cite{mercier2021deep} & 12.00 & 4.00  & 21.88 & 0.00 \\
     \midrule
     Ours & \textbf{68.00} & \textbf{34.00} & \textbf{84.38} & \textbf{37.50}\\
     \bottomrule
    \end{tabular}
    \caption{\textbf{Results on non-rigid object localization.}}
    \label{tab:nonrigid}
\end{table}

\section{Conclusion}
This paper presents a general one-shot object localization algorithm called OneLoc. It predicts pointwise center offset direction and relative bounding-box size and then uses a voting scheme to generate the final bounding box. Experiments show that the proposed method achieves the state-of-the-art performance on OnePose and LINEMOD datasets. Besides, it can also be applied to one-shot multi-instance detection and non-rigid object localization.

\section{Acknowledgements}
This work was supported by the National Key Research and Development Project of China (No. 2021ZD0110700), the National Natural Science Foundation of China under Grant 51975350, Shanghai Municipal Science and Technology Major Project (2021SHZDZX0102), Shanghai Qi Zhi Institute, and SHEITC (2018-RGZN-02046). This work was also supported by the  Shanghai AI development project (2020-RGZN-02006) and ``cross research fund for translational medicine'' of Shanghai Jiao Tong University (zh2018qnb17, zh2018qna37, YG2022ZD018).





{\small
\bibliographystyle{ieee_fullname}
\bibliography{egbib}
}

\end{document}